\documentclass[conference,a4paper]{IEEEtran}
\IEEEoverridecommandlockouts
\usepackage[hidelinks]{hyperref}
\usepackage{cite}
\usepackage{amsmath,amssymb,amsfonts}
\usepackage{algorithmic}
\usepackage{graphicx}
\usepackage{textcomp}
\usepackage{xcolor}
\usepackage{amsfonts}
\usepackage{booktabs}
\usepackage{siunitx}
\def\BibTeX{{\rm B\kern-.05em{\sc i\kern-.025em b}\kern-.08em
    T\kern-.1667em\lower.7ex\hbox{E}\kern-.125emX}}
    
\newcommand*{\doi}[1]{\href{http://dx.doi.org/#1}{doi: #1}}

\begin{document}

\title{Spatial-Temporal Graph Attention Fuser for Calibration in IoT Air Pollution Monitoring Systems\\

}
\author{
    \IEEEauthorblockN{Keivan Faghih Niresi\textsuperscript{*}, Mengjie Zhao\textsuperscript{*}, Hugo Bissig\textsuperscript{\textdagger}, Henri Baumann\textsuperscript{\textdagger},  Olga Fink\textsuperscript{*}}
    \IEEEauthorblockA{\textsuperscript{*}Intelligent Maintenance and Operations Systems (IMOS) Lab, EPFL, Switzerland \\
    \textsuperscript{\textdagger}Federal Institute of Metrology (METAS), Switzerland \\
    \{keivan.faghihniresi, mengjie.zhao, olga.fink\}@epfl.ch}\{henri.baumann, hugo.bissig\}@metas.ch
    \thanks{Accepted to ``IEEE SENSORS 2023". © 2023 IEEE.  Personal use of this material is permitted.  Permission from IEEE must be obtained for all other uses, in any current or future media, including reprinting/republishing this material for advertising or promotional purposes, creating new collective works, for resale or redistribution to servers or lists, or reuse of any copyrighted component of this work in other works.}
    }
    


\maketitle

\begin{abstract}
The use of Internet of Things (IoT) sensors for air pollution monitoring has significantly increased, resulting in the deployment of low-cost sensors. Despite this advancement, accurately calibrating these sensors in uncontrolled environmental conditions remains a challenge. To address this, we propose a novel approach that leverages  graph neural networks, specifically the graph attention network module, to enhance the calibration process by fusing data from sensor arrays. Through our experiments, we demonstrate the effectiveness of our approach in significantly improving the calibration accuracy of sensors in IoT air pollution monitoring platforms.
\end{abstract}


\section{Introduction}
With the growth of densely populated cities, increased traffic, and pollution from energy production and industrial activities, air pollution has emerged as a pressing concern, exerting a negative impact on both the environment and human health. Various substances such as tropospheric ozone (O\textsubscript 3), nitrogen dioxide (NO\textsubscript 2) and carbon monoxide (CO) are sources of air pollution\cite{spinelle2015field}. Among these pollutants, tropospheric ozone stands out as a particularly detrimental factor. Hence, accurate monitoring of air quality, particularly ozone levels, is crucial for effective pollution management and public health protection\cite{lu2019meteorology}.

Air pollution monitoring platforms have emerged as a result of the increasing utilization of the Internet of Things (IoT), enabling the collection of real-time data from multiple  sensors placed in various locations. However, the accuracy of the ozone sensors built into these IoT-based solutions is a concern \cite{lewis2018low}. Manufacturers often neglect the calibration process  and overlook how environmental conditions, such as air temperature and humidity can impact sensor performance. To enhance the reliability and functionality of gas sensors, air quality monitoring platforms commonly incorporate low-cost temperature and relative humidity sensors in addition  to gas sensors \cite{cross2017use}. This redundancy in measuring pollutant concentration helps improve the calibration quality of gas sensors.

Ferrer-Cid \emph{et al}.\ \cite{pau1} studied the application of multi-sensor data fusion techniques using machine learning and weighted averaging methods in air pollution monitoring platforms. Their study demonstrated the effectiveness of traditional machine learning algorithms, including support vector regression (SVR), random forest (RF), and k-nearest neighbors (KNN), in fusing sensor data and performing calibration compared to weighted averaging. However, it was observed that the performance of these models heavily relies on feature engineering (e.g., applying partial least squares (PLS) regression to mitigate multicollinearity), emphasizing the importance of carefully selecting and engineering the relevant features such as applying fast Fourier transform\cite{marathe2021currentsense}. Additionally, these methods usually neglect the temporal and spatial interdependencies present in the sensor readings, which could potentially provide valuable information for further improving calibration accuracy \cite{allka}.

\begin{figure*}
  \centering
\includegraphics[width=0.95\linewidth]{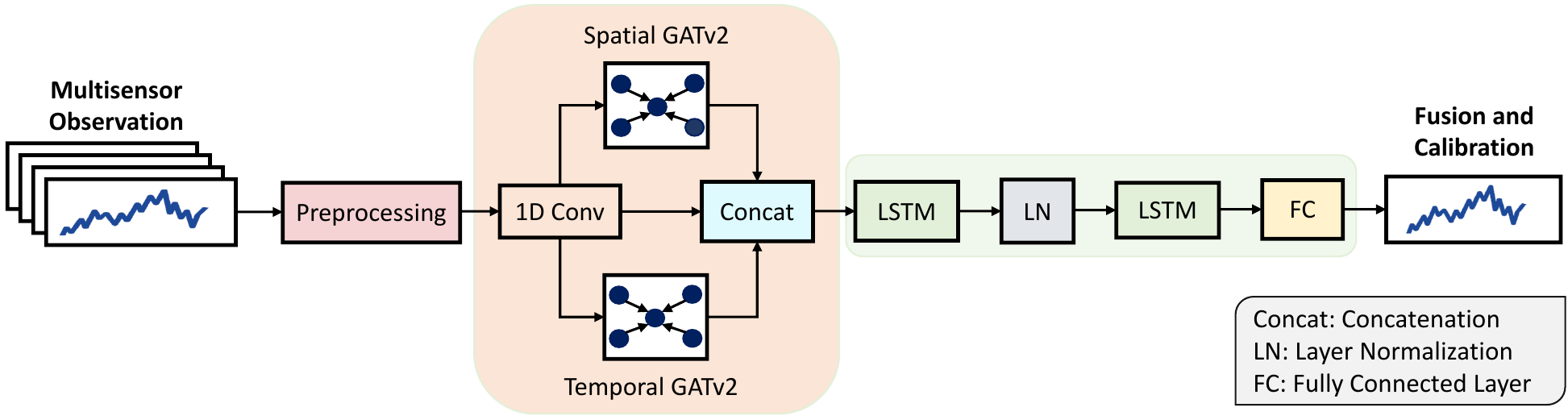}
  \caption{The overall architecture of STGAT-Fuser is designed to capture both spatial and temporal correlations for data fusion and calibration. It consists of multiple key components, including a 1D convolution module, a temporal and spatial Graph Attention Network (GATv2) module, an LSTM module, layer normalization (LN) and fully connected layers.}
  \label{fig:architecture}
\end{figure*}

Graph neural networks (GNNs)  have proven effective in capturing spatial-temporal interrelationships in data, not only in cases where a graph structure is already present but also in scenarios involving spatially distributed data subject to spatial and temporal correlations \cite{Scarselli,Wu, zhou2020graph}.  As a result, they offer significant potential in diverse domains, including traffic data analysis\cite{Cui2020graph,jiang2022graph, bui2022spatial}, recommendation systems\cite{rec}, and biological networks\cite{biological}. However, despite their wide-ranging applications, there exists a notable research gap regarding  the utilization of GNNs in the context of IoT air pollution monitoring platforms. This gap presents an exciting opportunity to explore the potential advantages and challenges associated with leveraging the spatial and temporal relationships in sensor networks.

In this paper, we propose a GNN-based model to address the challenge of a low-cost multisensor fusion for calibrating ozone sensors in IoT air pollution monitoring platforms. GNNs have the ability to effectively capture sensor interdependencies, allowing them to learn representations that incorporate information from neighboring nodes and the overall graph topology. Our approach particularly leverages the Graph Attention Network (GAT)\cite{veličković2018graph} into the calibration process to enhance data fusion and improve the accuracy of ozone sensor calibration.

The main objective of this research is to overcome the limitations of conventional machine learning-based calibration methods by utilizing the benefits of the GAT. By fusing data from multiple sensors and effectively capturing the underlying spatial-temporal relationships, our approach provides  a cost-effective solution for calibrating ozone sensors in IoT air pollution monitoring platforms. The proposed method has the potential to improve the reliability and accuracy of ozone measurements, thereby enabling more effective air quality management strategies.

The remainder of this paper is organized as follows. Section~\ref{sec:method} outlines the methodology, including the details of the GAT and the calibration process. In Section~\ref{sec:results}, experimental results and performance evaluation are presented. Finally, Section~\ref{sec:conclusion} concludes the paper.

\section{Proposed Methods}
\label{sec:method}

\subsection{Graph Attention Networks (GATs)}
The incorporation of attention mechanisms in GNNs has led to substantial progress in the field by enhancing representation learning with graphs through self-attention mechanisms. The fundamental operation in GAT is the aggregation of neighboring node features through the attention mechanism\cite{veličković2018graph}. However, it has been observed that the initial attention mechanism used in GATs, commonly referred to as ``static" attention, has limitations in terms of its expressive power. This form of attention remains static,  meaning that the ranking of attention scores is independent of the query node's characteristics. While this restricted form of attention, although effective in certain scenarios, it hinders GATs from addressing more complex graph problems. To overcome this limitation, a more expressive variant called GATv2 has been proposed\cite{brody2022how}. GATv2 incorporates a dynamic attention mechanism that enables nodes to adaptively adjust their attention based on their own features and interactions with neighboring nodes. In the following, we review the mathematics behind the attention mechanism in GNNs. 

Let $\mathcal{G} = (\mathcal{V}, \mathcal{E})$ be a graph where $\mathcal{V}$ is the set of nodes, and $\mathcal{E}$ is the set of edges. Each node $v \in \mathcal{V}$ is associated with a feature vector $\mathbf{h}_v \in \mathbb{R}^d$, where $d$ represents the dimensionality of the node features. In GATv2, for each node $i$ with its neighbors $\mathcal{N}(i)$, an aggregation operation is performed to derive a new node representation $\mathbf{h}^{\prime}_i$, which is formulated as follows:
\begin{equation}
   \mathbf{h}^{\prime}_i = \sigma \left( \sum_{j \in \mathcal{N}(i)} \alpha_{ij} \mathbf{W} \mathbf{h}_j \right),
\end{equation}
where
$\mathbf{W}$ is a weight matrix,
$\mathbf{h}_j$ is the feature vector of node $j$,
$\alpha_{ij}$ is the attention coefficient between node $i$ and node $j$, and
$\sigma$ is a non-linear activation function, such as LeakyReLU.
The attention coefficient $\alpha_{ij}$ is computed by:
\begin{equation}
    \alpha_{ij} = \text{softmax}_j \left( e_{ij} \right),
\end{equation}
where $e_{ij}$ is the unnormalized attention score between node $i$ and node $j$, computed by a shared attention mechanism $\mathbf{a}$, which  is typically a single-layer feedforward neural network. The design of $e_{ij}$ in GATv2 addresses  the ``static" attention  limitation  in the original GAT by modifying the operators in $e_{ij}$ with the following formulation:
\begin{equation}
    e(\mathbf{h}_i, \mathbf{h}_j) = \mathbf{a}^\textnormal{T}\text{LeakyReLU}\big(\mathbf{W}\cdot \big[\mathbf{h}_i || \mathbf{h}_j \big]\big),
\end{equation}
where $||$ represents the concatenation operator.

\subsection{Overall Architecture}
After performing data preprocessing (Section \ref{sec:results}-A), the input data is initially processed through a 1D convolutional (1D Conv) layer to extract relevant features. Building upon the successful application of temporal and spatial graph attention layers as powerful feature extractors for multivariate time series in \cite{MTADGAT}, we propose a novel model, named Spatial-Temporal Graph Attention Fuser (STGAT-Fuser), that incorporates these attention layers to capture long-term dependencies, temporal relationships, and correlations among different sensor inputs. In addition, the proposed architecture is augmented with two long short-term memory (LSTM) layers \cite{lstm} and one layer normalization (LN) \cite{ba2016layer} applied between them, enabling the modeling of sequential dependencies and capturing long-term contextual information. This incorporation is beneficial for sensor fusion and calibration tasks as it allows the system to consider and leverage the extended historical context, thereby enhancing the understanding of complex temporal patterns and improving the accuracy and reliability of the fusion and calibration processes. To further process and calibrate the fused representation, a fully connected layer is employed. The comprehensive architecture of STGAT-Fuser (Fig. \ref{fig:architecture}) offers an effective solution for sensor fusion and calibration, making it a powerful tool for processing sequential data in sensor networks. During the training process, the mean squared error (MSE) loss function and the Adam optimizer \cite{kingma} with the learning rate of 0.001 are utilized.

\section{Experimental Results}
\label{sec:results}
\subsection{Dataset and Preprcoessing}
In order to evaluate the effectiveness of the proposed method, we conducted experiments using a real-world dataset obtained from low-cost sensors deployed in IoT platforms. Specifically, we utilized an hourly-sampled  dataset (from June 2017 to October 2017) which was collected as part of the H2020 CAPTOR project. This dataset includes ozone (O\textsubscript3) measurements captured by metal–oxide (MOX) and electrochemical (EC)-based sensor technologies\cite{barcelo2021h2020, ripoll2019testing}. For reference, the sensors were deployed in close proximity to reference stations operated and monitored by governmental organizations. These reference stations are equipped with highly accurate instruments and serve as ground truth information for the measurements captured by IoT platforms. Our analysis focused on data collected from a specific node (C-17017, R69-17) located in Tona, Spain due to the availability of a comparatively large number samples from this particular location. This node comprised four MOX sensors, one EC sensor, as well as an air temperature sensor and an air relative humidity sensor.

The dataset was partitioned based on the chronological order of the data for training and evaluation. The initial 80\% of the data was used for training, while the remaining portion was evenly split into validation and testing sets. This approach preserves the temporal sequence of the data, enabling the model to be trained and evaluated on diverse time periods, ensuring its generalizability and robustness. To prevent overfitting during training part, early stopping was applied. This technique monitors the error on the validation set and terminates the training process if there is no improvement in the validation set error for 40 consecutive epochs. To address the issue of varying ranges of different features and prevent bias during training  min-max scaling was applied. This scaling technique normalizes the measurements of each sensor by transforming them to a range of $[0, 1]$  using the minimum and maximum values from the training dataset.
To meet the requirements of convolutional neural networks (CNN), LSTM, and STGAT-Fuser models, the dataset is preprocessed using a sliding time window approach, with a window size of 4 and a stride size of 1. 


\subsection{Performance Evaluation}
In this study, we compared the performance of STGAT-Fuser  to several models, including multiple linear regression (MLR), SVR, multilayer perceptron (MLP), CNN, and LSTM. The comparison was based on the evaluation metrics of root mean squared error (RMSE) and mean absolute error (MAE). Table~\ref{tab:tab1} presents the results of each method. It is worth mentioning that for CNN, MLP, LSTM, and STGAT-Fuser, the mean and standard deviation values obtained from five independent runs are reported. Our findings reveal that MLR exhibits the poorest performance, indicating that the linear modeling approach fails to effectively capture the complex interactions among sensors. SVR demonstrates improved performance compared to MLR, yet it falls short of achieving highly competitive results. MLP, LSTM, and CNN models demonstrate similar performance levels, while STGAT-Fuser surpasses all other models in terms of both RMSE and MAE, indicating its superior performance in learning the spatial an temporal relationships and dependencies in the sensor data. The ablation study results in Table \ref{tab:tab2} highlight the crucial role of both Spatial and Temporal GATv2 modules in the calibration and fusion process, as indicated by the lower performance of the stacked CNN-LSTM (without both GATv2) configuration compared to other ablation studies.

\begin{table}
    \begin{center}
    \caption{Comparison of RMSE and MAE Metrics for Sensor Fusion Calibration using Proposed and Baseline Approaches.  Mean and standard Deviation Results with MLP, CNN, LSTM, and STGAT-Fuser Models Over Five Runs.}
    \label{tab:tab1}
\begin{tabular}{ccc} \toprule
    {Methods} & {RMSE ($\mu gr / m^3$)} & {MAE ($\mu gr / m^3$)} \\ \midrule
    MLR  & 8.190 & 7.065  \\
    SVR  & 6.088  & 4.868   \\
    MLP  & 5.527 $\pm$ 0.24  & 4.219 $\pm$ 0.18 \\
    CNN  & 5.411 $\pm$ 0.30  & 4.183 $\pm$ 0.20\\ 
    LSTM  & 5.422 $\pm$ 0.32  & 4.187 $\pm$ 0.25 \\
    STGAT-Fuser  & \textbf{5.197 $\pm$ 0.28}  & \textbf{4.076 $\pm$ 0.21}  \\ \bottomrule
\end{tabular}
\end{center}
\end{table}

\begin{table}
    \begin{center}
    \caption{Ablation Study for STGAT-Fuser Architecture}
    \label{tab:tab2}
\begin{tabular}{ccc} \toprule
    {Methods} & {RMSE ($\mu gr / m^3$)} & {MAE ($\mu gr / m^3$)} \\ \midrule
    w/o Temporal GATv2  & 5.237 $\pm$ 0.28 & 4.109 $\pm$ 0.26  \\
    w/o Spatial GATv2  & 5.278 $\pm$ 0.24 & 4.138 $\pm$ 0.25   \\
    w/o Both GATv2  & 5.290 $\pm$ 0.28  & 4.171 $\pm$ 0.21 \\ \midrule
    \textbf{STGAT-Fuser}  & \textbf{5.197 $\pm$ 0.28}  & \textbf{4.076 $\pm$ 0.21} 
    \\
        \bottomrule
\end{tabular}
\end{center}
\end{table}

\section{Conclusion}
\label{sec:conclusion}
This paper proposes an approach for multisensor fusion and calibration in IoT air pollution monitoring systems. By incorporating graph attention networks, CNN, and LSTM, the proposed method successfully captures the spatial and temporal relationships among sensors, resulting in improved calibration accuracy. This is  demonstrated with a real-world dataset. Future research directions may include further investigation of different graph neural network architectures and exploring the application of the proposed method in other domains. Overall, the findings presented in this paper contribute to the advancement of sensor fusion techniques and calibration methods, enhancing the performance of IoT air pollution monitoring systems.

\section*{Acknowledgment}
This research was funded by the Swiss Federal Institute of Metrology (METAS).

\bibliographystyle{IEEEtran}
\bibliography{IEEEabrv,Ref}

\end{document}